\documentclass[11pt,a4paper]{article}
\usepackage[hyperref]{emnlp2020}
\usepackage{times}
\usepackage{latexsym}

\usepackage{amsfonts}
\usepackage{multirow}
\usepackage{amsmath}

\usepackage{microtype}

\aclfinalcopy 
\def\aclpaperid{1504} 

\usepackage{graphicx}

\title{Dynamic Context Selection for Document-level Neural Machine Translation via Reinforcement Learning}

\author{
	Xiaomian Kang$^{1,2}$,
	Yang Zhao$^{1,2}$, 
	Jiajun Zhang$^{1,2,3}$, \and
	Chengqing Zong$^{1,2,4}$
	\\ 
	$^1$National Laboratory of Pattern Recognition, Institute of Automation, CAS, Beijing, China\\
	$^2$School of Artificial Intelligence, University of Chinese Academy of Sciences, Beijing, China\\
	$^3$Beijing Academy of Artificial Intelligence, Beijing, China\\
	$^4$CAS Center for Excellence in Brain Science and Intelligence Technology, Beijing, China\\
	{\tt \{xiaomian.kang, yang.zhao, jjzhang, cqzong\}@nlpr.ia.ac.cn}
}

\date{}

\begin{document}
	\maketitle
	\begin{abstract}
		Document-level neural machine translation has yielded attractive improvements. However, majority of existing methods roughly use all context sentences in a fixed scope. They neglect the fact that different source sentences need different sizes of context. To address this problem, we propose an effective approach to select dynamic context so that the document-level translation model can utilize the more useful selected context sentences to produce better translations. Specifically, we introduce a selection module that is independent of the translation module to score each candidate context sentence. Then, we propose two strategies to explicitly select a variable number of context sentences and feed them into the translation module. We train the two modules end-to-end via reinforcement learning. A novel reward is proposed to encourage the selection and utilization of dynamic context sentences. Experiments demonstrate that our approach can select adaptive context sentences for different source sentences, and significantly improves the performance of document-level translation methods.
	\end{abstract}

	\section{Introduction}
	
	Although neural machine translation (NMT) has achieved great progress in recent years \cite{cholearning,bahdanau2015neural,luong2015,vaswani2017attention}, when fed an entire document, standard NMT systems translate sentences in isolation without considering the cross-sentence dependencies. Consequently, document-level neural machine translation (DocNMT) methods are proposed to utilize source-side or target-side inter-sentence contextual information to improve translation quality over sentences in a document \cite{jean2017does,wang2017exploiting,tiedemann-scherrer-2017-neural,tu2018learning,kuang2017modeling,junczys-dowmunt-2019-microsoft,ma-etal-2020-simple}.
	
	More recently, researchers of DocNMT mainly focus on exploring various attention-based networks to leverage the cross-sentence context efficiently, and evaluate the special discourse phenomena \cite{bawden2017evaluating,muller-etal-2018-large,voita2019,jwalapuram-etal-2019-evaluating}. However, there is still an issue that has received less attention: {\it which context sentences should be used when translating a source sentence?}
	
	\begin{table}[t!]
		\small
		\begin{center}
			\begin{tabular}{clcc}
				\hline
				\bf \# & \bf Test Context Settings & \bf Model1 & \bf Model2 \\
				\hline
				1 & previous 2 sentences & \underline{20.84} & 20.94 \\
				2 & previous 6 sentences & 20.90 & \underline{21.15} \\
				3 & select 2 from previous 6 & 22.03 & 22.14 \\
				4 & dynamic size from previous 6 & 22.90 & 22.74 \\
				\hline
				
			\end{tabular}
			\caption{The BLEU (\%) scores with different context settings. ``Model1'' and ``Model2'' are trained with previous 2 and 6 context sentences, respectively. Underlined results indicate that training and test context settings are consistent.}
			\label{f1}
		\end{center}
	\end{table}

	We conduct an experiment to verify an intuition: the translation of different source sentences requires different context. As shown in Table \ref{f1}, we train two DocNMT models and test them using various context settings\footnote{We apply a typical DocNMT method \cite{zhang2018improving} to train models on Zh$\rightarrow$En TED, and select 1,000 sentences to test. The BLEU of sentence-level baseline is 20.06.}. During the test, we obtain dynamic context sentences that achieve the best BLEU scores by traversing all the context combinations for each source sentence. Compared with the fixed size context (row 1 and 2), dynamic context (row 3 and 4) can significantly improve translation quality. Although row 2 uses more context, redundant information may hurt the results. Experiments indicate that only the limited context sentences are really useful, and they change with source sentences.
	
	Majority of existing DocNMT models set the context size or scope to be fixed. They utilize all of the
	previous $k$ context sentences \cite{voita2018context,zhang2018improving,miculicich2018document,voita2019,yang2019,ijcai2020-544}, or the full context in the entire document \cite{maruf2017document,tan2019,xiong2018modeling,ijcai2020-0551}. As a result, the inadequacy or redundancy of contextual information is almost inevitable. From this viewpoint, \citet{maruf2019selective} propose a selective attention approach that uses the {\it sparsemax} function \cite{martins2016softmax} instead of the softmax to normalize the attention weights. The sparsemax assigns the low probability in softmax to zero so that the model can focus on the sentences with high probability. However, the learning of attention weights lacks guidance, and they cannot handle the situation where the source sentences achieve the best translation results without relying on any context, which happens in about 39.4\% of sentences in the experiment.
	
	To address the problem, we propose an effective approach to select contextual sentences {\bf dynamically} for each source sentence in the document-level translation. Specifically, we propose a {\it Context Scorer} to score each candidate context sentence according to the currently translated source sentence. Then, we utilize two selection strategies to select useful context sentences for the translation module. The size of selected context is variable for different sentences. A core challenge of our approach is that the selection process is non-differentiable. Therefore, we leverage the reinforcement learning (RL) method to train the selection and DocNMT modules together. We design a novel reward to encourage the model to be aware of different context sentences and select more appropriate context to improve translation quality.
	
	In this paper, we make the following contributions:
	
	\begin{itemize}
		\item Our approach can measure the contribution of each context sentence to the source, and select dynamic context for the translation of different source sentences. Independent of the translation network, our approach is easily adaptable to existing DocNMT models. 
		
		\item We bridge the training of context selection and context-aware translation via reinforcement learning. Experiments show that our approach can significantly improve the performance of DocNMT models with the selected dynamic context sentences.
	\end{itemize}
	\section{Document-level Machine Translation}
	
	A standard DocNMT system generally translates a source sentence $X=\left\{ x_{1},\cdots,x_{I} \right\}$ to a target sentence $Y=\left \{ y_{1},\cdots,y_{T} \right \}$ with the aid of contextual information $\mathcal{Z}$ that is usually a subset of the candidate context set $\mathbb{Z}$. The model is trained to minimize the negative log-likelihood as:
	\begin{equation}
	\small
	\setlength\abovedisplayskip{0.1cm}
	\mathcal{L}_{mle} = - \sum_{t=1}^{T} \textrm{log} P\left( y_{t}\mid y_{< t},X,\mathcal{Z};\theta \right) \label{e1}
	\end{equation}
	
	Different granularity (word or sentence) and different sources (source-side or target-side) of contextual information $\mathcal{Z}$ have been explored. \citet{maruf2019selective} divide the candidate context set $\mathbb{Z}$ into two cases: {\it offline} where the context comes from the entire document, and {\it online} that only uses the past context. In this paper, we mainly focus on a general scenario, where DocNMT translates sentences with the online source-side context sentences.
	
	\begin{figure}[pt!]
		\centering
		\includegraphics[width=0.45\textwidth]{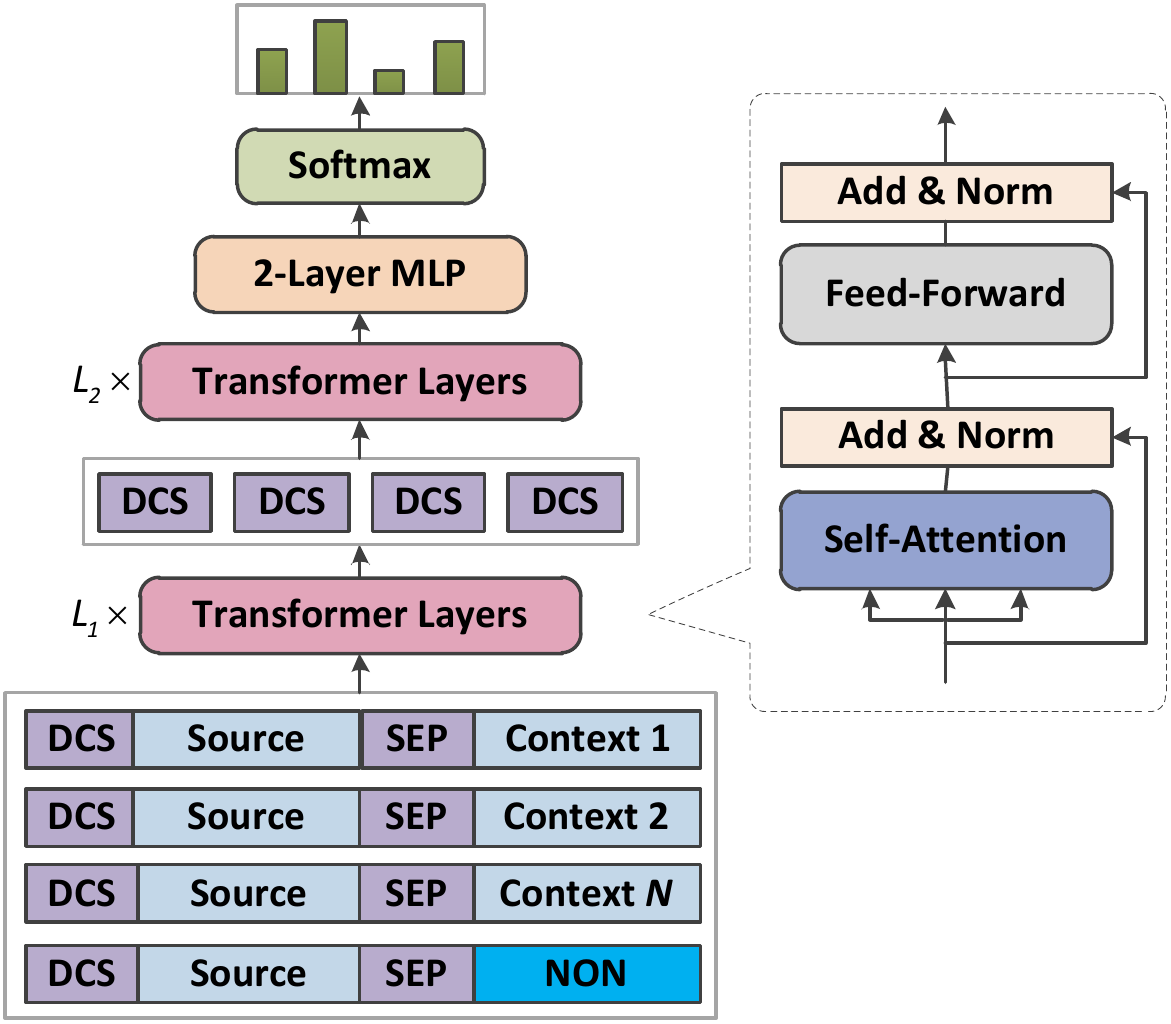}
		\caption{The architecture of context scorer. We add a special empty context sentence ``NON'' to help the decision of selection strategies. The details of Transformer layers are shown in the right dotted box.}
		\label{f1}
	\end{figure}
	
	\section{Dynamic Context Selection}
	
	Our approach translates a source sentence $X$ in the document in two steps. First, we select the appropriate context sentences for the translation of $X$ via the selection module. Independent of DocNMT module, this step is conducted before the context encoding in DocNMT module. The core component is a {\it Context Scorer} that calculates the contribution of each context sentence $z \in \mathbb{Z}$ to the translation of $X$ (sub-section \ref{s31}). According to the context scores, we propose two strategies to choose the useful context sentences (sub-section \ref{s32}). Second, we feed the selected context sentences into a DocNMT module to generate the translation.
	
	To overcome the non-differentiable behavior of the context selection and the lack of direct supervision when training the context scorer, we connect the two steps through the reinforcement learning strategy. We propose an effective reward that is related to the translation quality to guide the dynamic selection of context sentences and the optimization of parameters in DocNMT model (sub-section \ref{s33}).
	
	\subsection{Context Scorer}
	\label{s31}
	
	As Figure \ref{f1} shows, we obtain the representation of context sentences for scoring. Inspired by the popular pre-training language models such as GPT \cite{radford2018improving} and BERT \cite{devlin-etal-2019-bert}, we produce one instance by concatenating the source sentence with a context sentence, and adding a special symbol ``$\left \langle DCS \right \rangle$'' at the beginning and a separator token ``$\left \langle SEP \right \rangle$'' in between. The instance is fed into a stack of $L_1$ Transformer encoder layers. We believe the special symbol ``$\left \langle DCS \right \rangle$'' can encode the information of source-context sentence pairs well by the self-attention.
	
	For a candidate context sentence $z \in \mathbb{Z}$, its hidden state of ``$\left \langle DCS \right \rangle$'' after $L_1$ layers is extracted as the input to $L_2$ Transformer encoder layers to model the dependencies among context sentences. We denote the hidden state after $L_2$ layers as $h_{z} \in \mathbb{R}^{d_1}$. After that, we adopt a two-layer linear scorer network to measure the score as follows:
	\begin{equation}
	\small
	\label{e2}
	Score_{z} = \sigma \left( W_2 \left( W_1 h_{z} + b_1 \right) + b_2 \right)
	\end{equation}
	where $W_1 \in \mathbb{R}^{d_1 \times d_2}$, and $W_2 \in \mathbb{R}^{d_2 \times 1}$. $\sigma$ stands for the logistic sigmoid function.
	
	Considering the sampling operation during training process, we normalize all scores of context sentences in candidate set $\mathbb{Z}$ as a probability distribution: 
	\begin{equation}
	\small
	\label{e3}
	\mathcal{P}_{select} = \textrm{softmax} \left( \left[ Score_{1}; \cdots; Score_{ | \mathbb{Z} |  } \right] \right)
	\end{equation}
	where $\left[\cdot; \cdot\right]$ concatenates elements into a vector.
	
	\subsection{Selection Strategies}
	\label{s32}
	
	According to the selection probability in $\mathcal{P}_{select}$, we can obtain useful context sentences for the translation task. To select context dynamically, we add a special empty sentence ``$\left \langle NON \right \rangle$'' into the candidate context set, which stands for the situation that translates a source sentence without any context. As a result, we select those context sentences whose probability is higher than ``$\left \langle NON \right \rangle$''. If the probability of ``$\left \langle NON \right \rangle$'' is the highest, context size is zero. We call this strategy as {\bf probability-first}. The selected context sentences change dynamically with the change of source sentences, and the context size can range from 0 to $| \mathbb{Z} |$.
	
	In order to make a fair comparison with existing DocNMT models setting fixed context size, we also propose a {\bf size-first} strategy that selects the certain number of context sentences with the highest probability except ``$\left \langle NON \right \rangle$''. Despite of the fixed size, the context is still dynamic because selected sentences can be anywhere and discontinuous in the document.

	\subsection{Model Learning}
	\label{s33}
	
	\begin{figure}[pt!]
		\centering
		\includegraphics[width=0.46\textwidth]{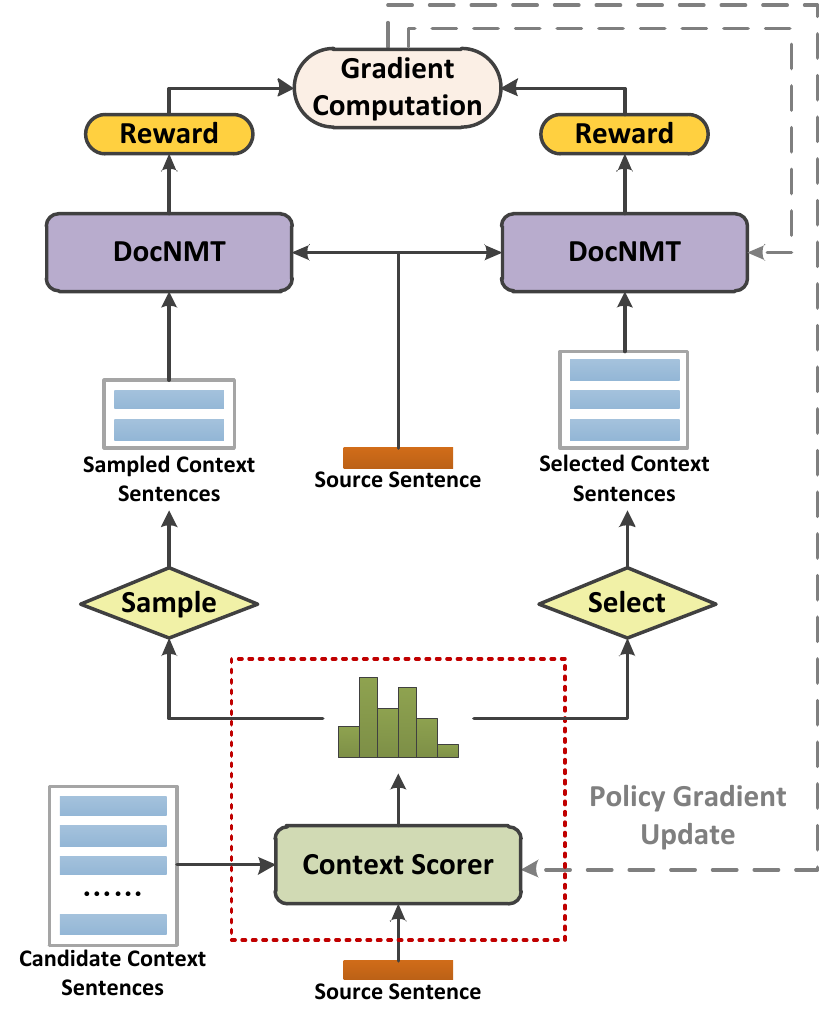}
		\caption{Reinforced training of the context selection and context-aware translation. The two DocNMT models share parameters.}
		\label{f2}
	\end{figure}
	
	Our strategies perform a non-differentiable hard selection, and it is difficult to decide which context sentences are helpful for the translation. It makes the training quite intractable. Therefore, we apply the policy gradient method to train the selection module and the document-level translation module in an end-to-end fashion through a novel reward. The reward encourages the model to select more useful context to improve the generation probability of the ground truth translations. Figure \ref{f2} shows the reinforcement-guided training process.
	
	\subsubsection{Modules Initialization}
	It is well known that a fine initialization of network is important to optimize the parameters in reinforcement learning.
	
	For DocNMT module that is usually trained in two stages \cite{tu2018learning,zhang2018improving,miculicich2018document,maruf2019selective}, we load the parameters of standard sentence-level NMT model to initialize the network.
	
	For the selection module, we simplify the initialization of context scorer as a binary classification task without considering the dependencies among context sentences. Its initialization contains two steps. First, we create pseudo labels for candidate context sentences. Each context sentence is labeled as \texttt{1} or \texttt{0}. The score in Eq. \ref{e2} is treated as the probability to predict label \texttt{1}. Specifically, pseudo labels are generated by an extra DocNMT model trained with a single random context sentence. We feed different candidate context sentences to the trained model to translate the same source sentence. Candidate context sentences with higher BLEU than ``$\left \langle NON \right \rangle$'' are labeled as \texttt{1}, while those with lower BLEU are labeled as \texttt{0}. Second, we train the context scorer to predict the pseudo labels. We share the parameters of embedding layer with initialized DocNMT model. The initial scorer is trained to minimize the cross-entropy loss.
	
	\subsubsection{Reward}
	Given that our goal of context selection is to improve translation quality, we propose a reward that can measure translation quality and is sensitive to the context changes \footnote{In our preliminary experiment, we try BLEU as reward but it is not sensitive enough to distinguish different context. Also, decoding a sequence to calculate BLEU is time-consuming.}.
	
	For a decoding time $t$, we calculate the cost of generating ground truth target word $y_t$ correctly as follows:
	\begin{equation}
	\small
	\label{e4}
	g_t = \textrm{log} P_{\tilde{y}_t^{\mathit{1st}}} - \textrm{log} P_{y_t} + \textrm{log}\left. P_{\tilde{y}_t^\mathit{1st}} \middle/ (P_{\tilde{y}_t^\mathit{1st}} - P_{\tilde{y}_t^\mathit{2nd}}) \right.
	\end{equation}
	where the first two items calculate the gap between the logarithmic probabilities of ground truth target word $y_t$ and the best word $\tilde{y}_t^{\mathit{1st}}$ whose probability is the top one in the prediction probability distribution. And the last item is a regularization that indicates the difference of probabilities between $\tilde{y}_t^{\mathit{1st}}$ and the word $\tilde{y}_t^{\mathit{2nd}}$ with the second-highest probability. The bigger difference means the higher confidence on the prediction.
	
	We obtain the average cost (whose value $>0$) of generating the ground truth sentence $Y=\left \{ y_{1},\cdots,y_{T} \right \}$, and utilize a monotone decreasing function to get the final reward bounded in $0 \sim 1$ as follows:
	\begin{equation}
	\small
	\label{e5}
	r (g) = {e}^{-g} = {e}^{-\frac{1}{T}\sum_{t=1}^{T} g_t}
	\end{equation}
	
	A high value of the reward means that it is easy to generate the ground truth. Therefore, the selected context sentences should be encouraged. Conversely, if a reward is low, generating the ground truth with the selected context would cost a lot, so the selection is discouraged.

	\subsubsection{Self-Critical Training}
	
	We train the whole model with the self-critical training method \cite{rennie2017self,bai-etal-2018-source}. The goal of RL training is to minimize the negative expected reward. And in practice, the loss is usually approximated with a single sample $u$ from the policy $\mathcal{P}$ as follows:
	\begin{equation}
	\small
	\mathcal{L}_{rl} = - \mathbb{E}_{u \sim \mathcal{P}} [ r(u) ] \approx - r(u), u \sim \mathcal{P}
	\label{e6}
	\end{equation}
	
	The self-critical training introduces a baseline reward $r(u')$ to reduce the variance of the gradient, where $u'$ is obtained by the inference algorithm at test time. The final gradient is estimated by:
	\begin{equation}
	\small
	\nabla \mathcal{L}_{rl} = (r(u) - r(u')) \nabla \textrm{log} P(u)
	\label{e7}
	\end{equation}
	
	Specifically, we denote the trainable parameters of the context scorer and DocNMT by $\omega$ and $\theta$, respectively. For each source sentence $X$, we select a set of context sentences $\mathcal{Z}^*$ by our selection strategies in sub-section \ref{s32}. Meanwhile, another set of context sentences $\hat{\mathcal{Z}}$ with the same size of $\mathcal{Z}^*$ is sampled according to $\mathcal{P}_{select}$ in equation \ref{e3}. Two sets of context sentences are fed into the same DocNMT module to obtain the rewards $r({\mathcal{Z}^*})$ and $r(\hat{\mathcal{Z}})$, respectively. Therefore, referring to equation \ref{e7}, the final gradient of the context scorer is calculated by:
	\begin{equation}
	\small
	\nabla_{\omega} \mathcal{L}(\omega ) = (r(\hat{\mathcal{Z}}) - r(\mathcal{Z}^*)) \nabla_{\omega} \textrm{log} P_{\omega}(\hat{\mathcal{Z}})
	\end{equation}
	where $P_{\omega}(\hat{\mathcal{Z}})$ is the probability of sampling $\hat{\mathcal{Z}}$ from $\mathcal{P}_{select}$. With the baseline reward $r(\mathcal{Z}^*)$ obtained by the current best policy (i.e., learned selection strategies), the method encourages model to explore more useful context (i.e., sampled context) that yields higher reward than the current best (i.e., selected context).
	
	For DocNMT module, we can combine the MLE objective (Eq. \ref{e1}) and RL objective (Eq. \ref{e6}) together to stabilize the training procedure \cite{wu-etal-2018-study} through a balance factor $\alpha$ as follows:
	\begin{equation}
	\small
	\mathcal{L} (\theta ) =  \alpha * \mathcal{L}_{mle} (Y \mid X, \hat{\mathcal{Z}}, \theta ) + (1 - \alpha) * \mathcal{L}_{rl} (\theta)
	\label{e9}
	\end{equation}
	
	We introduce the RL objective into DocNMT module so that the model can make better use of the selected context. The final RL gradient of DocNMT is calculated by:
	\begin{equation}
	\small
	\nabla_{\theta} \mathcal{L}_{rl} (\theta ) = (r(\hat{\mathcal{Z}}) - r(\mathcal{Z}^*)) \nabla_{\theta} \textrm{log} P_{\theta}(\hat{Y} \mid X, \hat{\mathcal{Z}})
	\label{e10}
	\end{equation}
	where $\hat{Y}$ is a sequence generated by current DocNMT model with the sampled context $\hat{\mathcal{Z}}$.
	
	\section{Experiment}
	
	\subsection{Datasets}
	We evaluate our approach on different domains of Chinese-English (Zh$\rightarrow$En) and English-German (En$\rightarrow$De) datasets. The corpora statistics are listed in Table \ref{t1}. For TED Talks in IWSLT17\footnote{\url{https://wit3.fbk.eu/mt.php?release=2017-01-trnted}}, we use \textit{dev{-}2010} as the development set, and \textit{tst{-}2010$\sim$2013} as the test set for both Zh$\rightarrow$En and En$\rightarrow$De language pairs. For News-Commentary v14\footnote{\url{http://data.statmt.org/news-commentary/v14}}, we use the \textit{newstest2017} for development and \textit{newstest2018} for testing. Europarl is a large scale corpus extracted from Europarl v7, and we use the same training, development and test sets as \citet{maruf2019selective}.
	
	\begin{table}[t!]
		\small
		\setlength{\abovecaptionskip}{0.3cm}
		\setlength{\belowcaptionskip}{-0.3cm}
		\begin{center}
			\begin{tabular}{llccccc}
				\hline
				\multicolumn{2}{c}{\bf Datasets} & \bf Training & \bf Dev & \bf Test \\
				\hline
				\multirow{2}{*}{Zh$\rightarrow$En} & TED & 0.23M & 0.88K & 4.68K \\
				& News & 0.31M & 2.00K & 3.98K \\
				\hline
				\multirow{3}{*}{En$\rightarrow$De} & TED & 0.21M & 0.89K & 4.70K \\
				& News & 0.33M & 3.00K & 3.00K \\
				& Europarl & 1.67M & 3.59K & 5.14K \\
				\hline
				
			\end{tabular}
			\caption{Dataset statistics in the number of sentences. }
			\label{t1}
		\end{center}
	\end{table}
	
	\subsection{Models}
	
	We compare our approach with the following methods: {\bf 1) {\scshape SentNmt}} \cite{vaswani2017attention} is a standard sentence-level Transformer model using the “base” version parameters. {\bf 2) TDNMT} \cite{zhang2018improving} introduces the contextual information by adding attention sub-layers at each encoder and decoder layer. We use 2 previous consecutive context sentences as they suggested. {\bf 3) HAN} \cite{miculicich2018document} uses 3 previous sentences as context. We adopt the “HAN encoder + HAN decoder” strategy that adds a hierarchical network on the top of the last encoder and decoder layer to model sentence-level and word-level contextual information. {\bf 4) SAN} \cite{maruf2019selective} utilizes all context in the entire document by calculating the sentence-level and word-level weights. It focuses on relevant context sentences through the sparsemax function. We choose the model that integrates the online context into encoder with “sparse-soft H-Attention”.
	
	We implement our approach and baseline methods based on the toolkit THUMT \cite{zhang2017thumt}. The parameters are the ``base'' version of the original Transformer \cite{vaswani2017attention}. The $d_1$ and $d_2$ in Eq. \ref{e2} are 512 and 256, respectively. We set the layers of $L_1=2$ and $L_2=2$. The effect of layer depth of context scorer and more implementation details are shown in the appendix.
	
	\begin{table*}[tbhp!]
		\small
		\setlength{\abovecaptionskip}{0.3cm}
		\setlength{\belowcaptionskip}{-0.3cm}
		\centering
		\begin{tabular}{clcccccccccc}
			\hline
			& & \multicolumn{3}{c}{\bf Context Settings } & \multicolumn{2}{c}{\bf TED} & \multicolumn{2}{c}{\bf News} & \bf Europarl \\
			\# & \bf Model & Scope & Size & Method & Zh-En & En-De & Zh-En & En-De & En-De \\
			\hline
			\multicolumn{10}{c}{\it Baselines with Fixed Context } \\
			\hline
			1 & {\scshape SentNmt} \cite{vaswani2017attention} & -- & -- & --  & 18.67 & 28.23 & 13.21 & 25.85 & 28.80 \\
			2 & HAN \cite{miculicich2018document} & 3 & 3 & full & 19.54 & 29.45 & 13.87 & 26.81 & 29.85 \\
			3 & TDNMT \cite{zhang2018improving} & 2 & 2 & full & 19.39 & 29.14 & 13.51 & 26.25 & 29.32 \\
			4 & HAN & 6 & 6 & full & 19.33 & 29.37 & 13.90 & 26.90 & 29.82 \\
			5 & TDNMT & 6 & 6 & full & 19.45 & 29.02 & 13.54 & 26.26 & 29.26 \\
			\hline
			\multicolumn{10}{c}{\it Baselines with Dynamic Context w/o RL Selection } \\
			\hline
			6 & HAN & 6 & 3 & random & 19.41 & 29.45 & 14.03 & 26.87 & 29.78 \\
			7 & TDNMT & 6 & 2 & random & 19.40 & 29.13 & 13.57 & 26.28 & 29.29 \\
			8 & SAN \cite{maruf2019selective} & all & dyn & attend & 19.60 & 29.41 & 14.08 & 26.79 & 29.81 \\
			9 & SAN & 6 & dyn & attend & 19.49 & 29.43 & 14.11 & 26.82 & 29.77 \\
			\hline
			\multicolumn{10}{c}{\it Our Methods} \\
			\hline
			10 & HAN + {\it DCS-SF} & 6 & 3 & select & 20.06 & 29.92 & 14.43 & 27.38 & 30.40 \\
			11 & TDNMT + {\it DCS-SF} & 6 & 2 & select & 20.09 & 29.70 & 14.36 & 26.93 & 29.89 \\
			12 & HAN + {\it DCS-PF}  & 3 & dyn & select & 19.97 & 29.87 & 14.37 & 27.34 & 30.36 \\
			13 & HAN +  {\it DCS-PF}  & 6 & dyn & select & 20.26 & \bf 30.22 & 14.48 & \bf 27.61 & \bf 30.48 \\
			14 & TDNMT + {\it DCS-PF}  & 2 & dyn & select & 19.91 & 29.50 & 14.19 & 26.62 & 29.64 \\
			15 & TDNMT + {\it DCS-PF}  & 6 & dyn & select & \bf 20.34 & 30.09 & 14.65 & 27.06 & 30.18 \\
			16 & SAN + {\it DCS-PF} & 6 & dyn & select & 20.18 & 30.13 & \bf 14.71 & 27.43 & 30.37 \\
			\hline

		\end{tabular}
		\caption{Performance of models on BLEU (\%) using different context settings. ``full'' means using all context in the scope. ``random'', ``attend'', and ``select'' stand for selecting sentences randomly, implicitly based on attention weights, and explicitly by our approaches, respectively. ``dyn'' stands for dynamic size. ``DCS-SF'' and ``DCS-PF'' mean dynamic context selection by size-first and probability-first strategies respectively. All results using ``DCS'' are statistically significantly (p-values $<$ 0.05) better than corresponding original DocNMT models.}
		\label{t3}
	\end{table*}

	\section{Results and Analysis}
	\subsection{Main Results}
	We use BLEU \cite{papineni2002bleu} score to evaluate the translation quality. Considering the memory limitation and complex sampling space, we select dynamic context from previous six sentences. Table \ref{t3} shows the performance of models utilizing different context settings. We always keep the same setting for training and test.
	
	{\bf Comparison with Fixed Context Methods.}  The performance of DocNMT models with fixed context is shown in row 2$\sim$5. Row 2 and 3 follow the context settings in the published papers. It can be found that using more context sentences indiscriminately (row 4 and 5) does not bring significant BLEU improvement. Instead, it increases computational cost.
	
	By contrast, our approach (row 10$\sim$15) can significantly improve translation quality on all datasets. Let us take the TDNMT models on Zh$\rightarrow$En TED for example. Row 11 applies the size-first strategy to select context sentences of the same size as original TDNMT model in row 3. The result achieves +0.70  BLEU improvement (20.09 vs. 19.39). Compared with row 5 that uses all context in previous six sentences, our approach can filter some redundant information and focus on fewer selected context sentences to gain +0.64 BLEU scores (20.09 vs. 19.45). On the other hand, even if the context is selected from previous two sentences (row 14), our model utilizing probability-first strategy can still improve original TDNMT by +0.52 BLEU (19.91 vs. 19.39). It indicates that useless context sentences still exist in a small scope. Conclusions are similar for other models and datasets.
	
	{\bf Comparison with Other Dynamic Methods.} Row 6$\sim$16 show the models trained and tested with dynamic context settings. Row 6 and 7 show a lower bound that randomly selects the fixed size context sentences. The results are similar to original models with the fixed size previous sentences (row 2 and 3). In contrast to the random selection, our approach (row 10 and 11) can select the same size of context sentences that are really helpful to generate better translations.
	
	SAN (row 8) implicitly selects context from all previous sentences through sharpening the attention weights. It resets low attention weights to zero to filter out some sentences. For a fair comparison, we also implement SAN in a limited context scope (row 9). Even if the candidate set is limited to previous six sentences, the BLEU does not decrease significantly. Different from SAN, our approach explicitly selects context sentences via reinforced guide. As row 16 shows, when added into SAN (row 9), our approach can obtain +0.69 BLEU gains (20.18 vs. 19.49) on Zh$\rightarrow$En TED by picking a more focused context candidate set for SAN. Furthermore, our approach can set the context size to be zero, but SAN cannot deal with this common cases that do not require any context.
	
	{\bf Comparison of Selection Strategies.} We also compare the two selection strategies proposed in section \ref{s32}. Results with probability-first strategy (row 13 and 15) are slightly better than size-first strategy (row 10 and 11). The size-first strategy has to contain some useless sentences because of the fixed size. By contrast, the probability-first strategy allows more flexible context selection of dynamic size. It can achieve +0.72 (20.26 vs. 19.54) and +0.95 (20.34 vs. 19.39) BLEU improvement on Zh$\rightarrow$En TED when applied to HAN and TDNMT model, respectively.

	\subsection{Effect of DocNMT Training}
	
	\begin{table}[t!]
		\small
		\setlength{\abovecaptionskip}{0.3cm}
		\setlength{\belowcaptionskip}{-0.5cm}
		\begin{center}
			\begin{tabular}{ccccc}
				\hline
				\multirow{2}{*}{\bf \#} & \multicolumn{2}{c}{\bf Training} & \multirow{2}{*}{\bf Balance RL Loss} & \multirow{2}{*}{\bf BLEU} \\
				& Scorer & DocNMT & & \\
				\hline
				\hline
				1 & $\times$ & $\times$ & -- & 29.04  \\
				2 & $\checkmark$ & $\times$ & -- & 29.58  \\
				\hline
				3 & $\checkmark$ & $\checkmark$ & $\alpha=1.00$ & 29.61  \\
				4 & $\checkmark$ & $\checkmark$ & $\alpha=0.75$ & 29.92 \\
				5 & $\checkmark$ & $\checkmark$ & $\alpha=0.50$ & 29.73  \\
				6 & $\checkmark$ & $\checkmark$ & $\alpha=0.25$ & 29.70  \\
				7 & $\checkmark$ & $\checkmark$ & $\alpha=0.00$ & 29.65  \\
				\hline
				
			\end{tabular}
			\caption{Effect of training settings for DocNMT models. BLEU scores are measured based on TDNMT on the development set in En$\rightarrow$De Europarl. $\checkmark$ means training the module while $\times$ means not. Row 1 stands for the original TDNMT model.}
			\label{t4}
		\end{center}
	\end{table}
	
	Our proposed context selection module is independent of the translation module. Therefore, the context scorer and DocNMT can be trained separately. As shown in Table \ref{t4}, we discuss the impact of selected context on the training of DocNMT. In row 2, we only train the context scorer, and keep the original DocNMT model unchanged as a component to calculate rewards. The result shows that our selection module can effectively distinguish between useful and useless context sentences for translation, and achieves +0.54 BLEU gains on the En$\rightarrow$De Europarl development set.
	
	We also explore whether the selected context would be helpful for the DocNMT training. We set the balance factor $\alpha$ Eq. \ref{e9} to be [0, 0.25, 0.5, 0.75, 1.0] in our experiments. Row 3 shows the model setting $\alpha=1.0$ that optimizes the standard MLE loss using the selected context sentences. Row 7 sets $\alpha=0.0$ to fine-tune DocNMT with the RL loss. By contrast, DocNMT models guided by the combination of MLE and RL loss can be learned better. We think the RL loss may make the model more sensitive to the selected context sentences. When $\alpha=0.75$, DocNMT can obtain the best BLEU score on development set, thus we use the setting in our experiments.
	
	\subsection{Distribution of Dynamic Context}
	
	\begin{figure}[pt!]
		\setlength{\abovecaptionskip}{0.3cm}
		\setlength{\belowcaptionskip}{-0.5cm}
		\centering
		\includegraphics[width=0.45\textwidth]{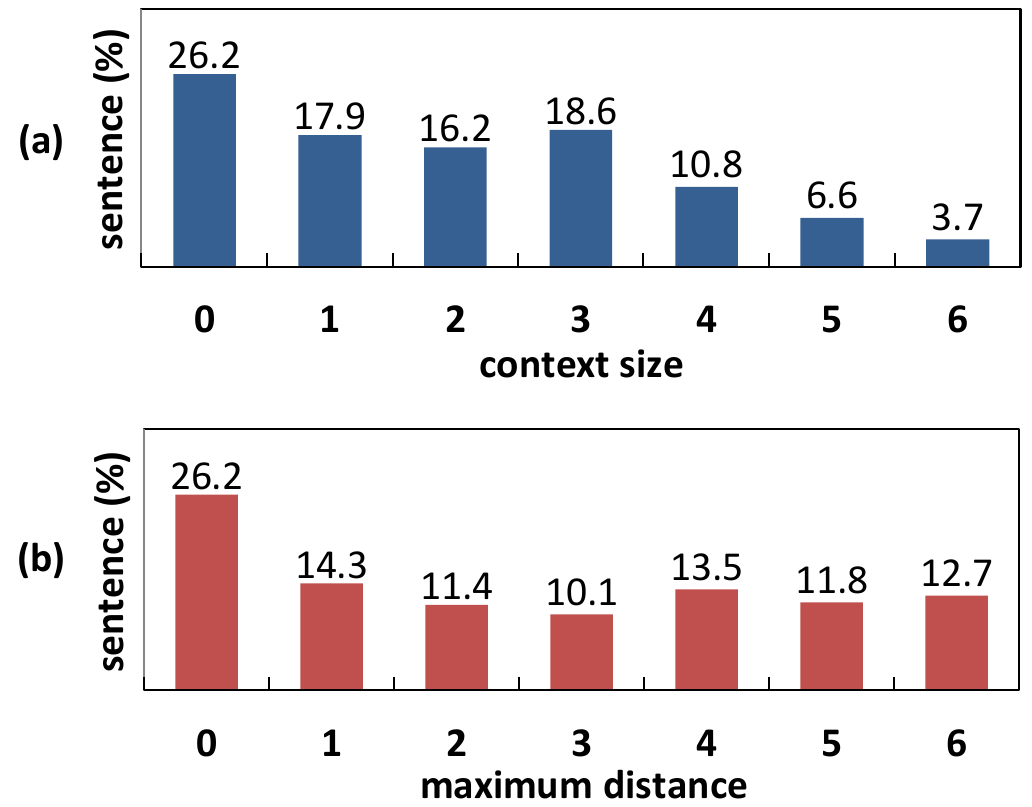}
		\caption{Distribution of dynamic context size and position. (a) shows the ratio of source sentences with different context sizes. (b) counts the maximum distance from the selected context sentences to their corresponding source sentences.}
		\label{f3}
	\end{figure}
	
	Figure \ref{f3} shows the distribution of different context sizes and maximum distances in the test sets of Zh$\rightarrow$En TED. Our approach selects context sentences whose size can range from zero to six. In Figure \ref{f3} (a), 78.9\% of source sentences tend to select no more than three context sentences. 26.2\% of sentences can be translated well without contextual information. The average context size over the test sets is 2.05 sentences. In Figure \ref{f3} (b), we show the maximum distances from the selected context sentences to the currently translated sentence. Except for the cases that need no context (distance 0), the distance distribution is relatively uniform. The total average distance is previous 2.57 sentences.
	
	\subsection{Selection of Empty Context}
	
	Our approach has the ability to select empty context for translation, which other models such as SAN \cite{maruf2019selective} cannot do. To evaluate whether the selected empty context is reasonable, we annotate a special test set that contains 500 sentences selected randomly from Zh$\rightarrow$En TED test sets. Each sentence is given its previous 6 sentences as context. Two annotators are instructed to mark context-empty sentences that can be translated well without any contextual information. The annotation details and statistics are shown in the appendix. The Cohen's Kappa value \cite{jacob1960} of annotation is 0.72. We gather sentences marked by both annotators as the final context-empty sentences (about 39.4\% in 500 sentences). Therefore, the test set is divided into context-empty and context-nonempty subsets. Their sizes are 197 and 303, respectively.
	
	\begin{table}[t!]
		\small
		\setlength{\belowcaptionskip}{-0.3cm}
		\begin{center}
			\begin{tabular}{ccc}
				\hline
				\bf Precision & \bf Recall & \bf F1 \\
				\hline
				68.46 & 51.78 & 58.96 \\
				\hline
				
			\end{tabular}
			\caption{Results of empty context prediction on 500 sentences with human annotation as reference.}
			\label{t5}
		\end{center}
	\end{table}
	
	\begin{table}[t!]
		\small
		\setlength{\belowcaptionskip}{-0.5cm}
		\begin{center}
			\begin{tabular}{lll}
				\hline
				\bf Model & \multicolumn{1}{c}{\bf Ctx-Empty} & \multicolumn{1}{c}{\bf Ctx-Nonempty} \\
				\hline
				{\scshape SentNmt} & 19.75$\qquad$ & 20.04$\qquad$ \\
				TDNMT & 20.49$\qquad$ & 20.72$\qquad$ \\
				\textit{\quad+DCS-PF} & 21.51 (+1.02) & 21.43 (+0.71) \\
				\hline
				
			\end{tabular}
			\caption{BLEU (\%) scores on the context-empty and context-nonempty test sets. ``+'' stands for the improvement when compared with TDNMT.}
			\label{t6}
		\end{center}
	\end{table}
	
	Table \ref{t5} shows the performance of our approach (using ``TDNMT+\textit{DCS-PF}'' model) for predicting empty context on the 500 annotated sentences. For the selection of empty context, our approach can achieve 58.96 F1-score.
	
	Table \ref{t6} shows the BLEU scores on the context-empty and context-nonempty subsets. Through our context selection, the improvement of BLEU on context-empty set is higher than context-nonempty set. The analysis indicates that our approach is aware of context-empty sentences, and can select empty context to improve translation quality.
	
	\subsection{Analysis of Discourse Phenomena}
	
	In addition to the selection of empty context, we also want to examine whether our approach can select context sentences that are helpful to improve the translation of discourse phenomena. 
	
	\citet{voita2019} construct contrastive test sets for English-Russian to evaluate four types of discourse phenomena (i.e., deixis, lexical cohesion, inflection and VP ellipses). Each test instance consists of a positive and several negative translations with incorrect phenomena. Models are evaluated by the accuracy that is defined as the proportion of times the generation probability of positive translation is higher than negative ones. Meanwhile, each instance has three context sentences. Among them, there is one and only one context sentence that is decisive in resolving the phenomena. It has been marked. Therefore, we can evaluate the accuracy of context selection, taking the marked context sentences as the standard answer.
	
	\begin{table}[t!]
		\small
		\setlength{\abovecaptionskip}{0.3cm}
		\setlength{\belowcaptionskip}{-0.6cm}
		\begin{center}
			\begin{tabular}{ccccc}
				\hline
				\bf Model & \bf deixis & \bf lex.c. & \bf ell.infl. & \bf ell.VP \\ 
				\hline
				\multirow{2}{*}{TDNMT+\textit{DCS-PF}} & 60.9 & 85.1 & 52.4 & 81.0 \\
				& 83.4 & 89.6 & 88.2 & 90.6 \\
				\hline
				\multirow{2}{*}{CADec+\textit{DCS-PF}} & 67.0 & 90.5 & 50.8 & 84.4 \\
				& 85.7 & 95.4 & 89.2 & 91.8 \\
				\hline
				
			\end{tabular}
			\caption{Accuracy (\%) of context selection on the discourse phenomena test sets. A model contains two rows: upper -- exact match, lower -- selected context contains the golden answer.}
			\label{t8}
		\end{center}
	\end{table}
	
	We use the same datasets as \citet{voita2019} to train models. Different from TDNMT \cite{zhang2018improving} that only uses source-side context, CADec \cite{voita2019} is proposed to utilize both source-side and target-side context. Based on CADec, we try to extend our approach in a simple way to select the target-side context. When the context scorer selects a source-side context sentence, the corresponding sentence-level translation is directly selected as target-side context.
	
	Table \ref{t8} shows the accuracy of context selection at four test sets. It can be found that our approach can select more than 85\% standard context sentences for special phenomena, and achieve more than 80\% exact match on lexical cohesion and VP ellipses sets.
	
	The accuracy of discourse phenomena are shown in Table \ref{t7}. TDNMT does not perform well because it only uses source-side context, which is unchanged in contrastive instances of test sets. Compared with original CADec, our approach can improve the performance of lexical cohesion. Although the simple way of selecting target-side context bears the risk of missing selection, the accuracy of some phenomena does not change significantly. Table \ref{t8} has shown that our approach can select useful target-side context in most cases. And the selection mechanism can make the model focus more on the useful context to resolve the discourse phenomena.
	
	\begin{table}[t!]
		\small
		\setlength{\abovecaptionskip}{0.3cm}
		\setlength{\belowcaptionskip}{-0.5cm}
		\begin{center}
			\begin{tabular}{lcccc}
				\hline
				\bf Model & \bf deixis & \bf lex.c. & \bf ell.infl. & \bf ell.VP \\
				\hline
				{\scshape SentNmt} & 50.0 & 45.9 & 53.0 & 28.4 \\
				TDNMT & 50.0 & 46.0 & 56.4 & 48.0 \\
				\hline
				CADec & 81.6 & 58.1 & 72.2 & 80.0 \\
				\textit{\quad+DCS-PF} & 79.2 & 62.0 & 71.8 & 80.8 \\
				\hline
			\end{tabular}
			\caption{Accuracy (\%) of discourse phenomena.
			}
			\label{t7}
		\end{center}
	\end{table}
	
	\section{Related Work}
	Standard neural machine translation methods usually focus on the sentence-level translation \cite{cholearning,bahdanau2015neural,zhang2015deep,luong2015,tu-etal-2016-modeling,zhang-zong-2016-exploiting,vaswani2017attention,wang-etal-2019-compact,zhou-etal-2019-synchronous,ijcai2020-0559}. As a contrast, document-level neural machine translation methods mainly pay attention to how to utilize the cross-sentence context. Researchers propose various context-aware networks to utilize contextual information to improve the performance of DocNMT models on the translation quality \cite{jean2017does,tu2018learning,kuang2017modeling} or discourse phenomena \cite{bawden2017evaluating,xiong2018modeling,voita2019,voita-etal-2019-context}. However, most methods roughly leverage all context sentences in a fixed size that is tuned on development sets  \cite{wang2017exploiting,miculicich2018document,zhang2018improving,yang2019,voita2018context,ijcai2020-544} , or full context in the entire document \cite{maruf2017document,tan2019,kang144,ijcai2020-0551}. They ignore the individualized needs for context when translating different source sentences.
	
	Some works have noticed that not all context is useful \cite{jean2019context,kim2019}. \citet{kimura-etal-2019-selecting} explore the context selection in the single-encoder framework \cite{tiedemann-scherrer-2017-neural}, and select context sentences that yield highest forced back-translation probability. However, the method cannot optimize DocNMT model at training phase, and requires back-translation model at inference phrase. \citet{maruf2019selective} sharpen the attention weights between the source and context sentences through the {\it sparsemax} function, and implicitly select context with high attention weights. Nevertheless, the method lacks direct supervision over context selection, and it cannot cover the situation where context is not needed. Inspired by the extractive-abstractive summarization \cite{chen-bansal-2018-fast}, our approach is different from above DocNMT methods. Our approach can explicitly select dynamic size (that can be 0) of context sentences for the translation of different source sentences.
	
	\section{Conclusion and Future Work}
	
	We propose a dynamic selection method to choose variable sizes of context sentences for document-level translation. The candidate context sentences are scored and selected by two proposed strategies. We train the whole model via reinforcement learning, and design a novel reward to encourage the selection of useful context sentences. When applied to existing DocNMT models, our approach can improve translation quality significantly. In the future, we will select context sentences in larger candidate space, and explore more effective ways to extend our approach to select target-side context sentences.
	
	\section*{Acknowledgments}
	
	We thank anonymous reviewers for their insightful comments and suggestions. The research work described in this paper has been supported by the Natural Science Foundation of China under Grant No. U1836221 and 61673380. The research work in this paper has also been supported by Beijing Advanced Innovation Center for Language Resources and Beijing Academy of Artificial Intelligence (BAAI2019QN0504).

	\bibliographystyle{acl_natbib}
	\bibliography{emnlp2020}

\begin{thebibliography}{48}
\expandafter\ifx\csname natexlab\endcsname\relax\def\natexlab#1{#1}\fi

\bibitem[{Bahdanau et~al.(2015)Bahdanau, Cho, and Bengio}]{bahdanau2015neural}
Dzmitry Bahdanau, Kyunghyun Cho, and Yoshua Bengio. 2015.
\newblock Neural machine translation by jointly learning to align and
  translate.
\newblock In \emph{Proceedings of ICLR}.

\bibitem[{Bai et~al.(2018)Bai, Zhou, Zhang, Zhao, Hwang, and
  Zong}]{bai-etal-2018-source}
He~Bai, Yu~Zhou, Jiajun Zhang, Liang Zhao, Mei-Yuh Hwang, and Chengqing Zong.
  2018.
\newblock \href {https://www.aclweb.org/anthology/C18-1305} {Source critical
  reinforcement learning for transferring spoken language understanding to a
  new language}.
\newblock In \emph{Proceedings of the 27th International Conference on
  Computational Linguistics}, pages 3597--3607, Santa Fe, New Mexico, USA.
  Association for Computational Linguistics.

\bibitem[{Bawden et~al.(2018)Bawden, Sennrich, Birch, and
  Haddow}]{bawden2017evaluating}
Rachel Bawden, Rico Sennrich, Alexandra Birch, and Barry Haddow. 2018.
\newblock \href {https://doi.org/10.18653/v1/N18-1118} {Evaluating discourse
  phenomena in neural machine translation}.
\newblock In \emph{Proceedings of the 2018 Conference of the North {A}merican
  Chapter of the Association for Computational Linguistics: Human Language
  Technologies, Volume 1 (Long Papers)}, pages 1304--1313, New Orleans,
  Louisiana. Association for Computational Linguistics.

\bibitem[{Chen and Bansal(2018)}]{chen-bansal-2018-fast}
Yen-Chun Chen and Mohit Bansal. 2018.
\newblock \href {https://doi.org/10.18653/v1/P18-1063} {Fast abstractive
  summarization with reinforce-selected sentence rewriting}.
\newblock In \emph{Proceedings of the 56th Annual Meeting of the Association
  for Computational Linguistics (Volume 1: Long Papers)}, pages 675--686,
  Melbourne, Australia. Association for Computational Linguistics.

\bibitem[{Cho et~al.(2014)Cho, van Merrienboer, Gulcehre, Bahdanau, Bougares,
  Schwenk, and Bengio}]{cholearning}
Kyunghyun Cho, Bart van Merrienboer, Caglar Gulcehre, Dzmitry Bahdanau, Fethi
  Bougares, Holger Schwenk, and Yoshua Bengio. 2014.
\newblock \href {https://doi.org/10.3115/v1/D14-1179} {Learning phrase
  representations using rnn encoder-decoder for statistical machine
  translation}.
\newblock In \emph{Proceedings of the 2014 Conference on Empirical Methods in
  Natural Language Processing ({EMNLP})}, pages 1724--1734, Doha, Qatar.
  Association for Computational Linguistics.

\bibitem[{Cohen(1960)}]{jacob1960}
Jacob Cohen. 1960.
\newblock A coefficient of agreement for nominal scales.
\newblock \emph{Educational and Psychological Measurement}.

\bibitem[{Devlin et~al.(2019)Devlin, Chang, Lee, and
  Toutanova}]{devlin-etal-2019-bert}
Jacob Devlin, Ming-Wei Chang, Kenton Lee, and Kristina Toutanova. 2019.
\newblock \href {https://doi.org/10.18653/v1/N19-1423} {{BERT}: Pre-training of
  deep bidirectional transformers for language understanding}.
\newblock In \emph{Proceedings of the 2019 Conference of the North {A}merican
  Chapter of the Association for Computational Linguistics: Human Language
  Technologies, Volume 1 (Long and Short Papers)}, pages 4171--4186,
  Minneapolis, Minnesota. Association for Computational Linguistics.

\bibitem[{Jean and Cho(2019)}]{jean2019context}
S{\'e}bastien Jean and Kyunghyun Cho. 2019.
\newblock Context-aware learning for neural machine translation.
\newblock \emph{arXiv preprint arXiv:1903.04715}.

\bibitem[{Jean et~al.(2017)Jean, Lauly, Firat, and Cho}]{jean2017does}
Sebastien Jean, Stanislas Lauly, Orhan Firat, and Kyunghyun Cho. 2017.
\newblock Does neural machine translation benefit from larger context?
\newblock \emph{arXiv preprint arXiv:1704.05135}.

\bibitem[{Junczys-Dowmunt(2019)}]{junczys-dowmunt-2019-microsoft}
Marcin Junczys-Dowmunt. 2019.
\newblock \href {https://doi.org/10.18653/v1/W19-5321} {{M}icrosoft translator
  at {WMT} 2019: Towards large-scale document-level neural machine
  translation}.
\newblock In \emph{Proceedings of the Fourth Conference on Machine Translation
  (Volume 2: Shared Task Papers, Day 1)}, pages 225--233, Florence, Italy.
  Association for Computational Linguistics.

\bibitem[{Jwalapuram et~al.(2019)Jwalapuram, Joty, Temnikova, and
  Nakov}]{jwalapuram-etal-2019-evaluating}
Prathyusha Jwalapuram, Shafiq Joty, Irina Temnikova, and Preslav Nakov. 2019.
\newblock \href {https://doi.org/10.18653/v1/D19-1294} {Evaluating pronominal
  anaphora in machine translation: An evaluation measure and a test suite}.
\newblock In \emph{Proceedings of the 2019 Conference on Empirical Methods in
  Natural Language Processing and the 9th International Joint Conference on
  Natural Language Processing (EMNLP-IJCNLP)}, pages 2964--2975, Hong Kong,
  China. Association for Computational Linguistics.

\bibitem[{Kang and Zong(2020)}]{kang144}
Xiaomian Kang and Chengqing Zong. 2020.
\newblock \href {https://doi.org/10.11959/j.issn.2096-6652.202016} {Fusion of
  discourse structural position encoding for neural machine translation}.
\newblock \emph{Chinese Journal of Intelligent Science and Technology},
  2(2):144--152.

\bibitem[{Kim et~al.(2019)Kim, Tran, and Ney}]{kim2019}
Yunsu Kim, Duc~Thanh Tran, and Hermann Ney. 2019.
\newblock \href {https://doi.org/10.18653/v1/D19-6503} {When and why is
  document-level context useful in neural machine translation?}
\newblock In \emph{Proceedings of the Fourth Workshop on Discourse in Machine
  Translation (DiscoMT 2019)}, pages 24--34, Hong Kong, China. Association for
  Computational Linguistics.

\bibitem[{Kimura et~al.(2019)Kimura, Iida, Cui, Hung, Utsuro, and
  Nagata}]{kimura-etal-2019-selecting}
Ryuichiro Kimura, Shohei Iida, Hongyi Cui, Po-Hsuan Hung, Takehito Utsuro, and
  Masaaki Nagata. 2019.
\newblock \href {https://www.aclweb.org/anthology/W19-6616} {Selecting
  informative context sentence by forced back-translation}.
\newblock In \emph{Proceedings of Machine Translation Summit XVII Volume 1:
  Research Track}, pages 162--171, Dublin, Ireland. European Association for
  Machine Translation.

\bibitem[{Kuang et~al.(2018)Kuang, Xiong, Luo, and Zhou}]{kuang2017modeling}
Shaohui Kuang, Deyi Xiong, Weihua Luo, and Guodong Zhou. 2018.
\newblock \href {https://www.aclweb.org/anthology/C18-1050} {Modeling coherence
  for neural machine translation with dynamic and topic caches}.
\newblock In \emph{Proceedings of the 27th International Conference on
  Computational Linguistics}, pages 596--606, Santa Fe, New Mexico, USA.
  Association for Computational Linguistics.

\bibitem[{Luong et~al.(2015)Luong, Pham, and Manning}]{luong2015}
Thang Luong, Hieu Pham, and Christopher~D. Manning. 2015.
\newblock \href {https://doi.org/10.18653/v1/D15-1166} {Effective approaches to
  attention-based neural machine translation}.
\newblock In \emph{Proceedings of the 2015 Conference on Empirical Methods in
  Natural Language Processing}, pages 1412--1421, Lisbon, Portugal. Association
  for Computational Linguistics.

\bibitem[{Ma et~al.(2020)Ma, Zhang, and Zhou}]{ma-etal-2020-simple}
Shuming Ma, Dongdong Zhang, and Ming Zhou. 2020.
\newblock \href {https://doi.org/10.18653/v1/2020.acl-main.321} {A simple and
  effective unified encoder for document-level machine translation}.
\newblock In \emph{Proceedings of the 58th Annual Meeting of the Association
  for Computational Linguistics}, pages 3505--3511, Online. Association for
  Computational Linguistics.

\bibitem[{Martins and Astudillo(2016)}]{martins2016softmax}
Andre Martins and Ramon Astudillo. 2016.
\newblock \href {http://proceedings.mlr.press/v48/martins16.html} {From softmax
  to sparsemax: A sparse model of attention and multi-label classification}.
\newblock volume~48 of \emph{Proceedings of Machine Learning Research}, pages
  1614--1623, New York, New York, USA. PMLR.

\bibitem[{Maruf and Haffari(2018)}]{maruf2017document}
Sameen Maruf and Gholamreza Haffari. 2018.
\newblock \href {https://doi.org/10.18653/v1/P18-1118} {Document context neural
  machine translation with memory networks}.
\newblock In \emph{Proceedings of the 56th Annual Meeting of the Association
  for Computational Linguistics (Volume 1: Long Papers)}, pages 1275--1284,
  Melbourne, Australia. Association for Computational Linguistics.

\bibitem[{Maruf et~al.(2019)Maruf, Martins, and Haffari}]{maruf2019selective}
Sameen Maruf, Andr{\'e}~FT Martins, and Gholamreza Haffari. 2019.
\newblock \href {https://doi.org/10.18653/v1/N19-1313} {Selective attention for
  context-aware neural machine translation}.
\newblock In \emph{Proceedings of the 2019 Conference of the North {A}merican
  Chapter of the Association for Computational Linguistics: Human Language
  Technologies, Volume 1 (Long and Short Papers)}, pages 3092--3102,
  Minneapolis, Minnesota. Association for Computational Linguistics.

\bibitem[{Miculicich et~al.(2018)Miculicich, Ram, Pappas, and
  Henderson}]{miculicich2018document}
Lesly Miculicich, Dhananjay Ram, Nikolaos Pappas, and James Henderson. 2018.
\newblock \href {https://doi.org/10.18653/v1/D18-1325} {Document-level neural
  machine translation with hierarchical attention networks}.
\newblock In \emph{Proceedings of the 2018 Conference on Empirical Methods in
  Natural Language Processing}, pages 2947--2954, Brussels, Belgium.
  Association for Computational Linguistics.

\bibitem[{M{\"u}ller et~al.(2018)M{\"u}ller, Rios, Voita, and
  Sennrich}]{muller-etal-2018-large}
Mathias M{\"u}ller, Annette Rios, Elena Voita, and Rico Sennrich. 2018.
\newblock \href {https://doi.org/10.18653/v1/W18-6307} {A large-scale test set
  for the evaluation of context-aware pronoun translation in neural machine
  translation}.
\newblock In \emph{Proceedings of the Third Conference on Machine Translation:
  Research Papers}, pages 61--72, Brussels, Belgium. Association for
  Computational Linguistics.

\bibitem[{Papineni et~al.(2002)Papineni, Roukos, Ward, and
  Zhu}]{papineni2002bleu}
Kishore Papineni, Salim Roukos, Todd Ward, and Wei-Jing Zhu. 2002.
\newblock \href {https://doi.org/10.3115/1073083.1073135} {Bleu: a method for
  automatic evaluation of machine translation}.
\newblock In \emph{Proceedings of the 40th Annual Meeting of the Association
  for Computational Linguistics}, pages 311--318, Philadelphia, Pennsylvania,
  USA. Association for Computational Linguistics.

\bibitem[{Radford et~al.(2018)Radford, Narasimhan, Salimans, and
  Sutskever}]{radford2018improving}
Alec Radford, Karthik Narasimhan, Tim Salimans, and Ilya Sutskever. 2018.
\newblock Improving language understanding by generative pre-training.

\bibitem[{Rennie et~al.(2017)Rennie, Marcheret, Mroueh, Ross, and
  Goel}]{rennie2017self}
Steven~J Rennie, Etienne Marcheret, Youssef Mroueh, Jerret Ross, and Vaibhava
  Goel. 2017.
\newblock Self-critical sequence training for image captioning.
\newblock In \emph{Proceedings of the IEEE Conference on Computer Vision and
  Pattern Recognition}, pages 7008--7024.

\bibitem[{Sennrich et~al.(2016)Sennrich, Haddow, and
  Birch}]{sennrich2015neural}
Rico Sennrich, Barry Haddow, and Alexandra Birch. 2016.
\newblock \href {https://doi.org/10.18653/v1/P16-1162} {Neural machine
  translation of rare words with subword units}.
\newblock In \emph{Proceedings of the 54th Annual Meeting of the Association
  for Computational Linguistics (Volume 1: Long Papers)}, pages 1715--1725,
  Berlin, Germany. Association for Computational Linguistics.

\bibitem[{Tan et~al.(2019)Tan, Zhang, Xiong, and Zhou}]{tan2019}
Xin Tan, Longyin Zhang, Deyi Xiong, and Guodong Zhou. 2019.
\newblock \href {https://doi.org/10.18653/v1/D19-1168} {Hierarchical modeling
  of global context for document-level neural machine translation}.
\newblock In \emph{Proceedings of the 2019 Conference on Empirical Methods in
  Natural Language Processing and the 9th International Joint Conference on
  Natural Language Processing (EMNLP-IJCNLP)}, pages 1576--1585, Hong Kong,
  China. Association for Computational Linguistics.

\bibitem[{Tiedemann and Scherrer(2017)}]{tiedemann-scherrer-2017-neural}
J{\"o}rg Tiedemann and Yves Scherrer. 2017.
\newblock \href {https://doi.org/10.18653/v1/W17-4811} {Neural machine
  translation with extended context}.
\newblock In \emph{Proceedings of the Third Workshop on Discourse in Machine
  Translation}, pages 82--92, Copenhagen, Denmark. Association for
  Computational Linguistics.

\bibitem[{Tu et~al.(2018)Tu, Liu, Shi, and Zhang}]{tu2018learning}
Zhaopeng Tu, Yang Liu, Shuming Shi, and Tong Zhang. 2018.
\newblock \href {https://doi.org/10.1162/tacl_a_00029} {Learning to remember
  translation history with a continuous cache}.
\newblock \emph{Transactions of the Association for Computational Linguistics},
  6:407--420.

\bibitem[{Tu et~al.(2016)Tu, Lu, Liu, Liu, and Li}]{tu-etal-2016-modeling}
Zhaopeng Tu, Zhengdong Lu, Yang Liu, Xiaohua Liu, and Hang Li. 2016.
\newblock \href {https://doi.org/10.18653/v1/P16-1008} {Modeling coverage for
  neural machine translation}.
\newblock In \emph{Proceedings of the 54th Annual Meeting of the Association
  for Computational Linguistics (Volume 1: Long Papers)}, pages 76--85, Berlin,
  Germany. Association for Computational Linguistics.

\bibitem[{Vaswani et~al.(2017)Vaswani, Shazeer, Parmar, Uszkoreit, Jones,
  Gomez, Kaiser, and Polosukhin}]{vaswani2017attention}
Ashish Vaswani, Noam Shazeer, Niki Parmar, Jakob Uszkoreit, Llion Jones,
  Aidan~N Gomez, {\L}ukasz Kaiser, and Illia Polosukhin. 2017.
\newblock \href
  {http://papers.nips.cc/paper/7181-attention-is-all-you-need.pdf} {Attention
  is all you need}.
\newblock In \emph{Advances in Neural Information Processing Systems 30}, pages
  5998--6008. Curran Associates, Inc.

\bibitem[{Voita et~al.(2019{\natexlab{a}})Voita, Sennrich, and
  Titov}]{voita-etal-2019-context}
Elena Voita, Rico Sennrich, and Ivan Titov. 2019{\natexlab{a}}.
\newblock \href {https://doi.org/10.18653/v1/D19-1081} {Context-aware
  monolingual repair for neural machine translation}.
\newblock In \emph{Proceedings of the 2019 Conference on Empirical Methods in
  Natural Language Processing and the 9th International Joint Conference on
  Natural Language Processing (EMNLP-IJCNLP)}, pages 877--886, Hong Kong,
  China. Association for Computational Linguistics.

\bibitem[{Voita et~al.(2019{\natexlab{b}})Voita, Sennrich, and
  Titov}]{voita2019}
Elena Voita, Rico Sennrich, and Ivan Titov. 2019{\natexlab{b}}.
\newblock \href {https://doi.org/10.18653/v1/P19-1116} {When a good translation
  is wrong in context: Context-aware machine translation improves on deixis,
  ellipsis, and lexical cohesion}.
\newblock In \emph{Proceedings of the 57th Annual Meeting of the Association
  for Computational Linguistics}, pages 1198--1212, Florence, Italy.
  Association for Computational Linguistics.

\bibitem[{Voita et~al.(2018)Voita, Serdyukov, Sennrich, and
  Titov}]{voita2018context}
Elena Voita, Pavel Serdyukov, Rico Sennrich, and Ivan Titov. 2018.
\newblock \href {https://doi.org/10.18653/v1/P18-1117} {Context-aware neural
  machine translation learns anaphora resolution}.
\newblock In \emph{Proceedings of the 56th Annual Meeting of the Association
  for Computational Linguistics (Volume 1: Long Papers)}, pages 1264--1274,
  Melbourne, Australia. Association for Computational Linguistics.

\bibitem[{Wang et~al.(2017)Wang, Tu, Way, and Liu}]{wang2017exploiting}
Longyue Wang, Zhaopeng Tu, Andy Way, and Qun Liu. 2017.
\newblock \href {https://doi.org/10.18653/v1/D17-1301} {Exploiting
  cross-sentence context for neural machine translation}.
\newblock In \emph{Proceedings of the 2017 Conference on Empirical Methods in
  Natural Language Processing}, pages 2826--2831, Copenhagen, Denmark.
  Association for Computational Linguistics.

\bibitem[{Wang et~al.(2019)Wang, Zhou, Zhang, Zhai, Xu, and
  Zong}]{wang-etal-2019-compact}
Yining Wang, Long Zhou, Jiajun Zhang, Feifei Zhai, Jingfang Xu, and Chengqing
  Zong. 2019.
\newblock \href {https://doi.org/10.18653/v1/P19-1117} {A compact and
  language-sensitive multilingual translation method}.
\newblock In \emph{Proceedings of the 57th Annual Meeting of the Association
  for Computational Linguistics}, pages 1213--1223, Florence, Italy.
  Association for Computational Linguistics.

\bibitem[{Wu et~al.(2018)Wu, Tian, Qin, Lai, and Liu}]{wu-etal-2018-study}
Lijun Wu, Fei Tian, Tao Qin, Jianhuang Lai, and Tie-Yan Liu. 2018.
\newblock \href {https://doi.org/10.18653/v1/D18-1397} {A study of
  reinforcement learning for neural machine translation}.
\newblock In \emph{Proceedings of the 2018 Conference on Empirical Methods in
  Natural Language Processing}, pages 3612--3621, Brussels, Belgium.
  Association for Computational Linguistics.

\bibitem[{Xiong et~al.(2019)Xiong, He, Wu, and Wang}]{xiong2018modeling}
Hao Xiong, Zhongjun He, Hua Wu, and Haifeng Wang. 2019.
\newblock \href
  {https://www.aaai.org/ojs/index.php/AAAI/article/view/4721/4599} {Modeling
  coherence for discourse neural machine translation}.
\newblock In \emph{Proceedings of the AAAI Conference on Artificial
  Intelligence}, volume~33, pages 7338--7345.

\bibitem[{Xu et~al.(2020)Xu, Xiong, van Genabith, and Liu}]{ijcai2020-544}
Hongfei Xu, Deyi Xiong, Josef van Genabith, and Qiuhui Liu. 2020.
\newblock \href {https://doi.org/10.24963/ijcai.2020/544} {Efficient
  context-aware neural machine translation with layer-wise weighting and
  input-aware gating}.
\newblock In \emph{Proceedings of the Twenty-Ninth International Joint
  Conference on Artificial Intelligence, {IJCAI-20}}, pages 3933--3940.
  International Joint Conferences on Artificial Intelligence Organization.

\bibitem[{Yang et~al.(2019)Yang, Zhang, Meng, Gu, Feng, and Zhou}]{yang2019}
Zhengxin Yang, Jinchao Zhang, Fandong Meng, Shuhao Gu, Yang Feng, and Jie Zhou.
  2019.
\newblock \href {https://doi.org/10.18653/v1/D19-1164} {Enhancing context
  modeling with a query-guided capsule network for document-level translation}.
\newblock In \emph{Proceedings of the 2019 Conference on Empirical Methods in
  Natural Language Processing and the 9th International Joint Conference on
  Natural Language Processing (EMNLP-IJCNLP)}, pages 1527--1537, Hong Kong,
  China. Association for Computational Linguistics.

\bibitem[{Zhang et~al.(2017)Zhang, Ding, Shen, Cheng, Sun, Luan, and
  Liu}]{zhang2017thumt}
Jiacheng Zhang, Yanzhuo Ding, Shiqi Shen, Yong Cheng, Maosong Sun, Huanbo Luan,
  and Yang Liu. 2017.
\newblock Thumt: An open source toolkit for neural machine translation.
\newblock \emph{arXiv preprint arXiv:1706.06415}.

\bibitem[{Zhang et~al.(2018)Zhang, Luan, Sun, Zhai, Xu, Zhang, and
  Liu}]{zhang2018improving}
Jiacheng Zhang, Huanbo Luan, Maosong Sun, FeiFei Zhai, Jingfang Xu, Min Zhang,
  and Yang Liu. 2018.
\newblock \href {https://doi.org/10.18653/v1/D18-1049} {Improving the
  transformer translation model with document-level context}.
\newblock In \emph{Proceedings of the 2018 Conference on Empirical Methods in
  Natural Language Processing}, pages 533--542, Brussels, Belgium. Association
  for Computational Linguistics.

\bibitem[{Zhang and Zong(2015)}]{zhang2015deep}
Jiajun Zhang and Chengqing Zong. 2015.
\newblock Deep neural networks in machine translation: An overview.
\newblock \emph{IEEE Intelligent Systems}, (5):16--25.

\bibitem[{Zhang and Zong(2016)}]{zhang-zong-2016-exploiting}
Jiajun Zhang and Chengqing Zong. 2016.
\newblock \href {https://doi.org/10.18653/v1/D16-1160} {Exploiting source-side
  monolingual data in neural machine translation}.
\newblock In \emph{Proceedings of the 2016 Conference on Empirical Methods in
  Natural Language Processing}, pages 1535--1545, Austin, Texas. Association
  for Computational Linguistics.

\bibitem[{Zhao et~al.(2020)Zhao, Zhang, Zhou, and Zong}]{ijcai2020-0559}
Yang Zhao, Jiajun Zhang, Yu~Zhou, and Chengqing Zong. 2020.
\newblock \href {https://doi.org/10.24963/ijcai.2020/559} {Knowledge graphs
  enhanced neural machine translation}.
\newblock In \emph{Proceedings of the Twenty-Ninth International Joint
  Conference on Artificial Intelligence, {IJCAI-20}}, pages 4039--4045.
  International Joint Conferences on Artificial Intelligence Organization.

\bibitem[{Zheng et~al.(2020)Zheng, Yue, Huang, Chen, and
  Birch}]{ijcai2020-0551}
Zaixiang Zheng, Xiang Yue, Shujian Huang, Jiajun Chen, and Alexandra Birch.
  2020.
\newblock \href {https://doi.org/10.24963/ijcai.2020/551} {Towards making the
  most of context in neural machine translation}.
\newblock In \emph{Proceedings of the Twenty-Ninth International Joint
  Conference on Artificial Intelligence, {IJCAI-20}}, pages 3983--3989.
  International Joint Conferences on Artificial Intelligence Organization.

\bibitem[{Zhou et~al.(2017)Zhou, Hu, Zhang, and Zong}]{zhou-etal-2017-neural}
Long Zhou, Wenpeng Hu, Jiajun Zhang, and Chengqing Zong. 2017.
\newblock \href {https://doi.org/10.18653/v1/P17-2060} {Neural system
  combination for machine translation}.
\newblock In \emph{Proceedings of the 55th Annual Meeting of the Association
  for Computational Linguistics (Volume 2: Short Papers)}, pages 378--384,
  Vancouver, Canada. Association for Computational Linguistics.

\bibitem[{Zhou et~al.(2019)Zhou, Zhang, and Zong}]{zhou-etal-2019-synchronous}
Long Zhou, Jiajun Zhang, and Chengqing Zong. 2019.
\newblock \href {https://doi.org/10.1162/tacl_a_00256} {Synchronous
  bidirectional neural machine translation}.
\newblock \emph{Transactions of the Association for Computational Linguistics},
  7:91--105.

\end{thebibliography}
	
	\appendix
	
	\section{Experimental Setup}
	
	\subsection{Parameters and Implementation}
	
	We implement all models based on the toolkit THUMT\footnote{\url{https://github.com/thumt/THUMT}} with the parameters of the “base” version of Transformer \cite{vaswani2017attention}. Specifically, we use 6 layers of encoder and decoder with 8 attention heads. The hidden size and feed-forward layer size are 512 and 2,048, respectively. For Zh$\rightarrow$En, Chinese and English vocabulary sizes are 30K and 25K, respectively. For En$\rightarrow$De, source-side and target-side share a vocabulary table. The vocabulary size is 30K. Chinese sentences are segmented into words by our in-house toolkit. English and German datasets are tokenized and truecased by the Moses toolkit\footnote{\url{https://github.com/moses-smt/mosesdecoder/tree/master/scripts}}. Words are segmented by byte-pair encoding \cite{sennrich2015neural}.
	
	We introduce a context scorer that is independent of the DocNMT models, which allows our approach to be easily deployed on many baseline DocNMT systems. Compared with original DocNMT models, the amount of additional parameters depends on the number of Transformer encoder layers $L_1$ and $L_2$ in the context scorer.
	
	\begin{table}[th]
		\small
		\begin{center}
			\begin{tabular}{cc|ccc}
				\hline
				\multicolumn{2}{c|}{\multirow{2}{*}{}} & \multicolumn{3}{c}{$L_1$} \\
				\multicolumn{2}{c|}{} & 1 & 2 & 4 \\
				\hline
				\multirow{3}{*}{$L_2$} & 0 & 29.35 & 29.58 & 29.64 \\
				& 1 & 29.60 & 29.72 & 29.77 \\
				& 2 & 29.76 & \bf 29.92 & 29.83 \\
				\hline
			\end{tabular}
			\caption{BLEU (\%) scores on En$\rightarrow$De TED development set using different layers of context scorer.}
			\label{t1}
		\end{center}
	\end{table}
	
	In Table \ref{t1}, we discuss the effect of layer depth of context scorer (defined in subsection 3.1). Experiments are conducted using ``TDNMT+\textit{DCS-PF}'' model with the balance factor $\alpha=0.75$. Our approach achieves the highest BLEU with a context scorer setting $L_1=2$ and $L_2=2$, which introduces 12.7M extra parameters to any original DocNMT models.
	
	\subsection{Training and Inference}
	
	For training, we use the Adam optimizer with $\beta_1=0.9$, $\beta_2=0.98$ and $\epsilon=10^{-9}$. We employ label smoothing with a value of 0.1 and dropout with a rate of 0.1. The batch size is 3,000 tokens. We employ 4 Titan Xp GPUs to train all models. Compared with original DocNMT (TDNMT), the training and testing speeds are slowed down by an order of 1.61 (mainly because of the generation of $\hat{Y}$ in Eq. \ref{e10}) and 1.05, respectively.
	
	We use \textit{multi-bleu.perl}\footnote{\url{https://github.com/moses-smt/mosesdecoder/blob/master/scripts/generic/multi-bleu.perl}} to compute the BLEU score. The beam size is set to 4. The significance test is conducted by the script “bootstraphypothesis-difference-significance.pl” in Moses.
	
	\section{Annotation and Statistics of Empty Context}
	
	In this section we describe the annotation process and statistics of the special test set constructed to evaluate the selection of empty context.
	
	\subsection{Annotation}
	
	We randomly select 500 sentences with previous 6 sentences as context from Chinese-English TED \textit{tst{-}2010$\sim$2013}. Each example to be annotated contains a source-reference sentence pair and six source-reference contextual sentence pairs. Two annotators proficient in both Chinese and English are instructed to annotate the sentences that can be translated well without any context. The process consists of three steps, and is carried out independently between two annotators.
	
	\textit{Step1.} Annotators are instructed to read a single source sentence $X$ without any context, and translate it into ${Y}'$ by themselves.
	
	\textit{Step2.} The reference $Y$ of the sentence $X$ is shown to annotators. Then, they are instructed to compare ${Y}'$ with $Y$ word-by-word to answer whether ${Y}'$ is appropriate.
	
	\textit{Step3.} Annotators are instructed to read source-reference contxtual sentences, and compare ${Y}'$ with $Y$ word-by-word again. After that, they are asked to determain whether ${Y}'$ needs to be modified better.
	
	If a annotator insists that his translation ${Y}'$ is appropriate at Step2 and needs no modification at Step3, the sentence $X$ is annotated as ``context-empty'', which means it can be translated well without relying on any context. Otherwise,  the sentence is annotated as ``context-nonempty''.
	
	\begin{table}[t!]
		\small
		\begin{center}
			\begin{tabular}{c|c|c|c}
				\hline
				\multicolumn{2}{c|}{\multirow{2}{*}{}} & \multicolumn{2}{c}{A1} \\
				\cline{3-4} \multicolumn{2}{c|}{} & Ctx-Empty & Ctx-Nonempty \\
				\hline
				\multirow{2}{*}{A2} & Ctx-Empty & 197 & 51 \\
				\cline{2-4} & Ctx-Nonempty & 20 & 232 \\
				\hline
			\end{tabular}
			\caption{Statistics of human annotation for empty context. A1 and A2 stand for two annotators.}
			\label{t3}
		\end{center}
	\end{table}
	
	Table \ref{t3} shows the statistics of annotation. The Cohen's Kappa value is 0.72. 197 context-empty sentences are annotated by both annotators. These sentences are gathered as the final context-empty test set. The other 303 sentences make up the context-nonempty test set.
	
	\subsection{Statistics of Empty Context Selection}
	
	Taking the human annotation in Table \ref{t3} as the golden test set, Table \ref{t4} shows the statistics of empty context prediction by our approach in subsection 5.3.
	
	\begin{table}[t!]
		\small
		\begin{center}
			\begin{tabular}{c|c|c|c}
				\hline
				\multicolumn{2}{c|}{\multirow{2}{*}{}} & \multicolumn{2}{c}{Human Annotation} \\
				\cline{3-4} \multicolumn{2}{c|}{} & Ctx-Empty & Ctx-Nonempty \\
				\hline
				\multirow{2}{*}{\textit{Ours}} & Ctx-Empty & 102 & 47 \\
				\cline{2-4} & Ctx-Nonempty & 95 & 256 \\
				\hline
			\end{tabular}
			\caption{Statistics of our approach (\textit{Ours}) for empty context prediction.}
			\label{t4}
		\end{center}
	\end{table}

\end{document}


\maketitle
\appendix

\section{Experimental Setup}

\subsection{Parameters and Implementation}

We implement all models based on the toolkit THUMT\footnote{\url{https://github.com/thumt/THUMT}} with the parameters of the “base” version of Transformer \cite{vaswani2017attention}. Specifically, we use 6 layers of encoder and decoder with 8 attention heads. The hidden size and feed-forward layer size are 512 and 2048, respectively. For Zh$\rightarrow$En, Chinese and English vocabulary sizes are 30K and 25K, respectively. For En$\rightarrow$De, source-side and target-side share a vocabulary table. The vocabulary size is 30K. Chinese sentences are segmented into words by our in-house toolkit. English and German datasets are tokenized and truecased by the Moses toolkit\footnote{\url{https://github.com/moses-smt/mosesdecoder/tree/master/scripts}}. Words are segmented by byte-pair encoding \cite{sennrich2015neural}.

Our approach introduces a context scorer that is independent of the DocNMT models, which allows our approach to be easily deployed on many baseline DocNMT systems. Compared with original DocNMT models, the amount of additional parameters depends on the number of self-attention layers in context scorer.

\begin{table}[th]
	\small
	\begin{center}
		\begin{tabular}{cc|ccc}
			\hline
			\multicolumn{2}{c|}{\multirow{2}{*}{}} & \multicolumn{3}{c}{$L_1$} \\
			\multicolumn{2}{c|}{} & 1 & 2 & 4 \\
			\hline
			\multirow{2}{*}{$L_2$} & 1 & 29.60 & 29.72 & 29.77 \\
			& 2 & 29.76 & \bf 29.92 & 29.83 \\
			\hline
		\end{tabular}
		\caption{BLEU (\%) scores on En$\rightarrow$De TED development set using different layers of context scorer.}
		\label{t1}
	\end{center}
\end{table}

In Table \ref{t1}, we discuss the effect of layer depth of context scorer (defined in subsection 3.1). Experiments are conducted using ``TDNMT+\textit{DCS-PF}'' model with the balance factor $\alpha=0.75$. Our approach achieves the highest BLEU with a context scorer setting $L_1=2$ and $L_2=2$, which introduces 12.7M parameters.

\subsection{Training and Inference}

For training, we use the Adam optimizer with $\beta_1=0.9$, $\beta_2=0.98$ and $\epsilon=10^{-9}$. We employ label smoothing with a value of 0.1 and dropout with a rate of 0.1. The batch size is 3,000 tokens. We employ 4 Titan Xp GPUs to train all models.

We use \textit{multi-bleu.perl}\footnote{\url{https://github.com/moses-smt/mosesdecoder/blob/master/scripts/generic/multi-bleu.perl}} to compute the BLEU score. The beam size is set to 4. The significance test is conducted by the script “bootstraphypothesis-difference-significance.pl” in Moses.

\section{Annotation and Statistics of Empty Context}

In this section we describe the annotation process and statistics of the special test set constructed to evaluate the selection of empty context.

\subsection{Annotation}

We randomly select 500 sentences with previous 6 sentences as context from Chinese-English TED \textit{tst{-}2010$\sim$2013}. Each example to be annotated contains a source-reference sentence pair and six source-reference contextual sentence pairs. Two annotators proficient in both Chinese and English are instructed to annotate the sentences that can be translated well without any context. The process consists of three steps, and is carried out independently between two annotators.

\textit{Step1.} Annotators are instructed to read a single source sentence $X$ without any context, and translate it into ${Y}'$ by themselves.

\textit{Step2.} The reference $Y$ of the sentence $X$ is shown to annotators. Then, they are instructed to compare ${Y}'$ with $Y$ word-by-word to answer whether ${Y}'$ is appropriate.

\textit{Step3.} Annotators are instructed to read source-reference contxtual sentences, and compare ${Y}'$ with $Y$ word-by-word again. After that, they are asked to determain whether ${Y}'$ needs to be modified better.

If a annotator insists that his translation ${Y}'$ is appropriate at step2 and still needs no modification at step3, the sentence $X$ is annotated as ``context-empty'', which means it can be translated well without relying any context. Otherwise,  the sentence is annotated as ``context-nonempty''.

\begin{table}[t!]
	\small
	\begin{center}
		\begin{tabular}{c|c|c|c}
			\hline
			\multicolumn{2}{c|}{\multirow{2}{*}{}} & \multicolumn{2}{c}{A1} \\
			\cline{3-4} \multicolumn{2}{c|}{} & Ctx-Empty & Ctx-Nonempty \\
			\hline
			\multirow{2}{*}{A2} & Ctx-Empty & 197 & 51 \\
			\cline{2-4} & Ctx-Nonempty & 20 & 232 \\
			\hline
		\end{tabular}
		\caption{Statistics of human annotation for empty context. A1 and A2 stand for two annotators.}
		\label{t3}
	\end{center}
\end{table}

Table \ref{t3} shows the statistics of annotation. The Cohen's Kappa value is 0.72. 197 context-empty sentences annotated by both annotators. They are gathered as the final context-empty test set. The other 303 sentences make up the context-nonempty test set.

\subsection{Statistics of Empty Context Selection}

Taking the human annotation in Table \ref{t3} as the golden test set, Table \ref{t4} shows the statistics of empty context prediction by our approach in subsection 5.3.

\begin{table}[t!]
	\small
	\begin{center}
		\begin{tabular}{c|c|c|c}
			\hline
			\multicolumn{2}{c|}{\multirow{2}{*}{}} & \multicolumn{2}{c}{Human Annotation} \\
			\cline{3-4} \multicolumn{2}{c|}{} & Ctx-Empty & Ctx-Nonempty \\
			\hline
			\multirow{2}{*}{\textit{Ours}} & Ctx-Empty & 102 & 47 \\
			\cline{2-4} & Ctx-Nonempty & 95 & 256 \\
			\hline
		\end{tabular}
		\caption{Statistics of our approach (\textit{Ours}) for empty context prediction.}
		\label{t4}
	\end{center}
\end{table}

\bibliographystyle{acl_natbib}
\bibliography{emnlp2020_app}